\documentclass[11pt, letterpaper]{article}
\usepackage[utf8]{inputenc} %
\usepackage[T1]{fontenc}    %
\usepackage[margin=1in]{geometry}

\usepackage{amsmath}
\usepackage{algorithm}
\usepackage{algpseudocode}
\usepackage{mathtools}
\usepackage{multirow}
\usepackage{subfig}
\usepackage{url}
\usepackage{hyperref}
\usepackage{amsfonts}
\usepackage{booktabs} 
\usepackage{amssymb}

\DeclareMathOperator*{\argmax}{arg\,max}
\DeclareMathOperator*{\argmin}{arg\,min}

\newcommand{\para}[1]{{\vspace{2pt} \noindent \textbf{#1}
    \hspace{6pt}}}
\newcommand{\etal}{{\em et al.\ }}
\newcommand{\eg}{{\em e.g.\ }}
\newcommand{\ie}{{\em i.e.\ }}
\newcommand{\etc}{{\em etc.}}

\newcommand\Alpha{\mathrm{A}}

\title{Learning to Counterfactually Explain Recommendations}
\author{Yuanshun Yao, Chong Wang\thanks{Work done while at ByteDance.}, Hang Li\\
ByteDance\\
\texttt{\{kevin.yao,chong.wang,lihang.lh\}@bytedance.com}
}
\date{}

\begin{document}

\maketitle

%%
%% The abstract is a short summary of the work to be presented in the
%% article.
\begin{abstract}
Recommender system practitioners are facing increasing pressure to explain recommendations. We explore how to explain recommendations using counterfactual logic, \ie ``Had you not interacted with the following items, we would not recommend it.'' Compared to the traditional explanation logic, counterfactual explanations are easier to understand, more technically verifiable, and more informative in terms of giving users control over recommendations. The major challenge of generating such explanations is the computational cost because it requires repeatedly retraining the models to obtain the effect on a recommendation caused by the absence of user history. We propose a learning-based framework to generate counterfactual explanations. The key idea is to train a \textit{surrogate model} to learn \textit{the effect of removing a subset of user history on the recommendation}. To this end, we first artificially simulate the counterfactual outcomes on the recommendation after deleting subsets of history. Then we train a surrogate model to learn the mapping between a history deletion and the corresponding change of the recommendation caused by the deletion. Finally, to generate an explanation, we find the history subset predicted by the surrogate model that is most likely to remove the recommendation. Through offline experiments and online user studies, we show our method, compared to baselines, can generate explanations that are more counterfactually valid and more satisfactory considered by users.
\end{abstract}

\section{Introduction}
\label{sec:intro}
Today's recommender systems need to fulfill a wide range of needs from users. One increasingly important demand is to explain the recommendation results that users receive. In the literature on explainable recommender system, one of the most popular approaches is \textit{item-based collaborative filtering}, \ie ``You receive the recommendation of A because you visited item B, C, and D.'' It explains the recommended item by showing a set of items similar to it. The logic is simple and easy to deploy.

% different approaches and we focus on item-based collaborative filtering logic because (1) Simple and easy to implement, (2) Easy to be understood by users, (3) Independent of recommender models (just finding relevant item).

However, there are three limitations in the item-based collaborative filtering logic. \textit{First}, the causal link is missing, \ie how did visiting B, C, and D lead to the recommendation of A? Without a more explicitly stated logic, users might not be satisfied. \textit{Second}, the technical claim is more or less questionable. Did the model truly make the decision \textit{solely} based on the item relevance? This is unlikely to be true in today's complex recommender system. Therefore the explanations might lead to doubts from the users or even potential legal accountability. \textit{Third}, this explanation does not give users any information about how to change the recommendations they receive-if they dislike them-by behaving differently in the future.

% problem of traditional logic: (1) Causal link missing: How did watching those movies lead to the recommendation? -> Users are not satisfied. (2) Questionable technical claim: Did the model really make the decision solely based on the movie relevance, in today’s complex recommender system? -> Doubt from the public and legal liability.

\begin{figure}[!t]
\centering
\begin{minipage}{.69\textwidth}
  \centering
   \includegraphics[width=\linewidth]{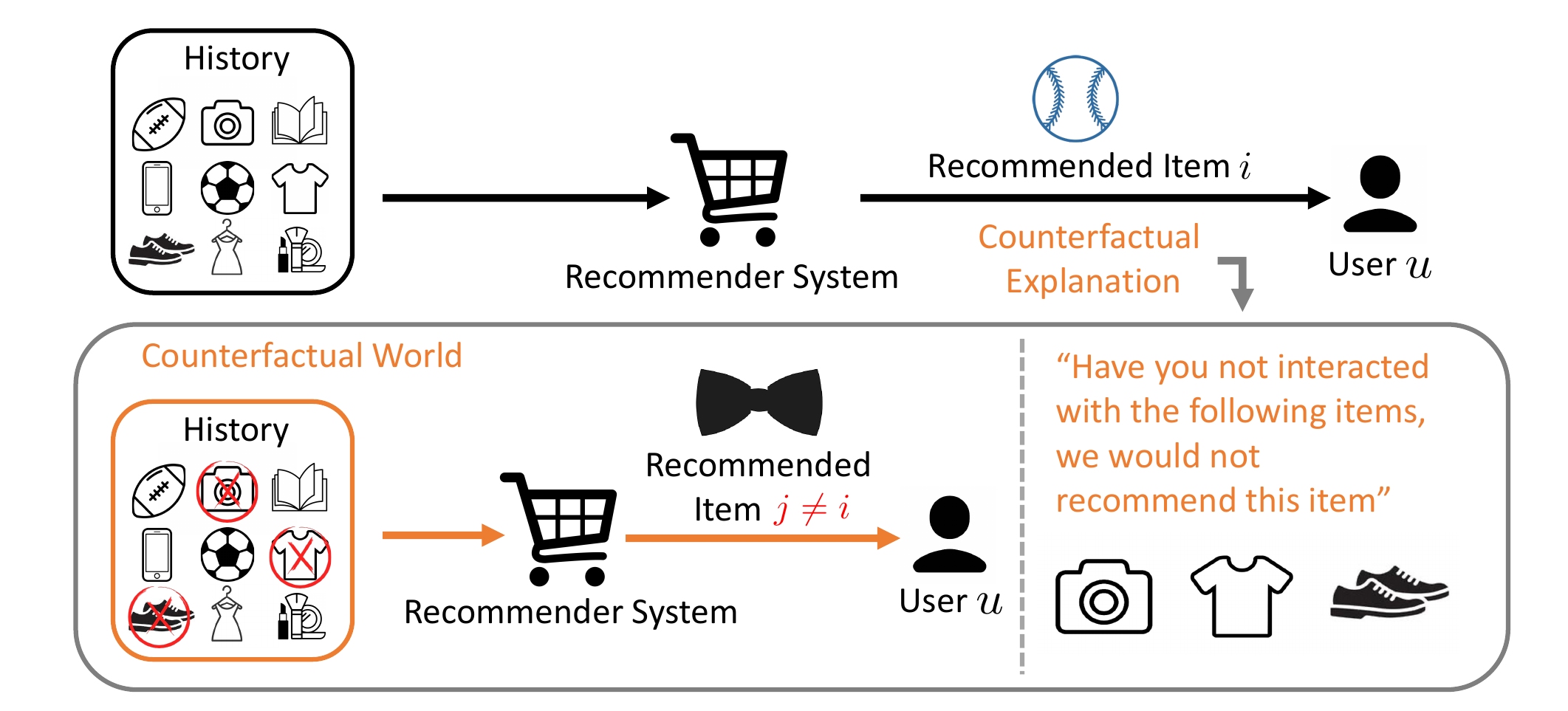}
  \caption{Example of a counterfactual explanation in a recommender system. When the system recommends an item $i$ to a user $u$, the explanation is ``Have you not interacted with the following items, we would not recommend this item.'' In other words, we show the items in user $u$'s interaction history that, if were removed, would change the current recommendation to a different one.}
  % \vspace{-0.5cm}
  \label{fig:exp_def}
\end{minipage}%
\hfill
\begin{minipage}{.29\textwidth}
  \centering
  \includegraphics[width=\linewidth]{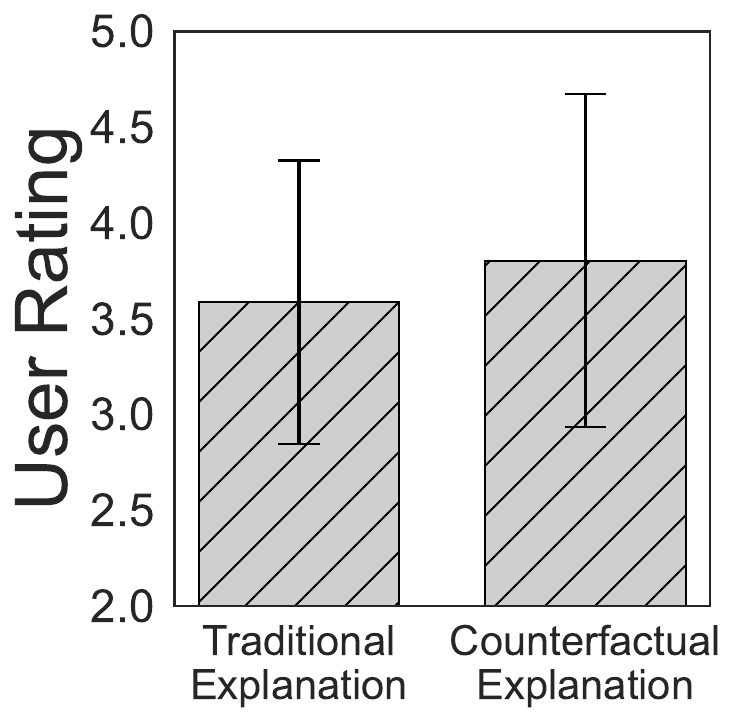}
    % \vspace{2cm}
  \captionof{figure}{Average user rating (1-5) on traditional (item-based collaborate-filtering) explanations and counterfactual explanations obtained from a user study in Section~\ref{sec:user}. Users consider counterfactual explanations to be more helpful.}
  \label{fig:user_results}
\end{minipage}
\end{figure}

One solution is to explain recommendations by \textit{counterfactual logic}, as illustrated in Figure~\ref{fig:exp_def}. Using the same example, the corresponding counterfactual explanation is ``Had you not visited item B, C, and D, we would not recommend A.''\footnote{Different from counterfactual examples in classification~\cite{wachter2017counterfactual}, we do not require the explanatory itemset size to be minimal. Instead we have a fixed size because displaying arbitrary number of items would 1) mess up the explanation user inferface and 2) overwhelm users if the list is too long. This is the same design choice in the conventional collaborative filtering explanations in recommender systems.} Compared to item-based collaborative filtering logic, it provides a more explicitly stated causal connection via ``What-if'' reasoning, and therefore more convincing, as shown in the psychology literature that counterfactual logic is a common way of thinking in our everyday life~\cite{kahneman1981simulation,mandel2005psychology,epstude2008functional,spellman1999possibility}. \textit{Additionally}, the claim is more technically verifiable because we can empirically test its correctness (by removing the claimed explanatory items and observing if the resulting recommendation actually changes or not). \textit{Finally}, if users do not like the current recommendation of A, by following the advice given by the explanation, \ie stopping interacting with items similar to B, C, D, then they can reduce the chance of receiving recommendations similar to A in the future, thus enabling users to have greater control over the recommendations. To demonstrate the advantage of counterfactual explanations, we later show through an user study (Section~\ref{sec:user}) that users consider counterfactual explanations to be more helpful in explaining recommendations than the traditional item-based collaborate filtering explanations (Figure~\ref{fig:user_results}).
% shows the average rating (ranging from 1 to 5, the higher is the better) given by the users.

% \begin{figure}[!t]
%   \centering
%   \includegraphics[width=0.8\linewidth]{figs/exp-def.pdf}
%   \caption{Example of a counterfactual explanation in a recommender system. When the system recommends an item $i$ to a user $u$, the explanation is ``Have you not interacted with the following items, we would not recommend this item.'' In other words, we show the set of items in user $u$'s interaction history that, if were removed, would change the current recommendation to a different item.}
%   % \vspace{-0.5cm}
%   \label{fig:exp_def}
% \end{figure}

Despite the benefits of counterfactual explanations, generating them is challenging. Unlike conventional counterfactual examples in the classification setting~\cite{wachter2017counterfactual} which assumes a \textit{fixed} model, in our case since user history is the recommendation model's training data, perturbing it would change the model weights while in the classification case, the model can remain fixed and we only need to perturb test samples. The naive way of generating such counterfactual explanations is computationally costly because it requires repeatedly retraining the recommender model after we delete training subsets of user history to observe the resulting change in the recommendation. 

\begin{figure}[!t]
  \centering
  \includegraphics[width=\linewidth]{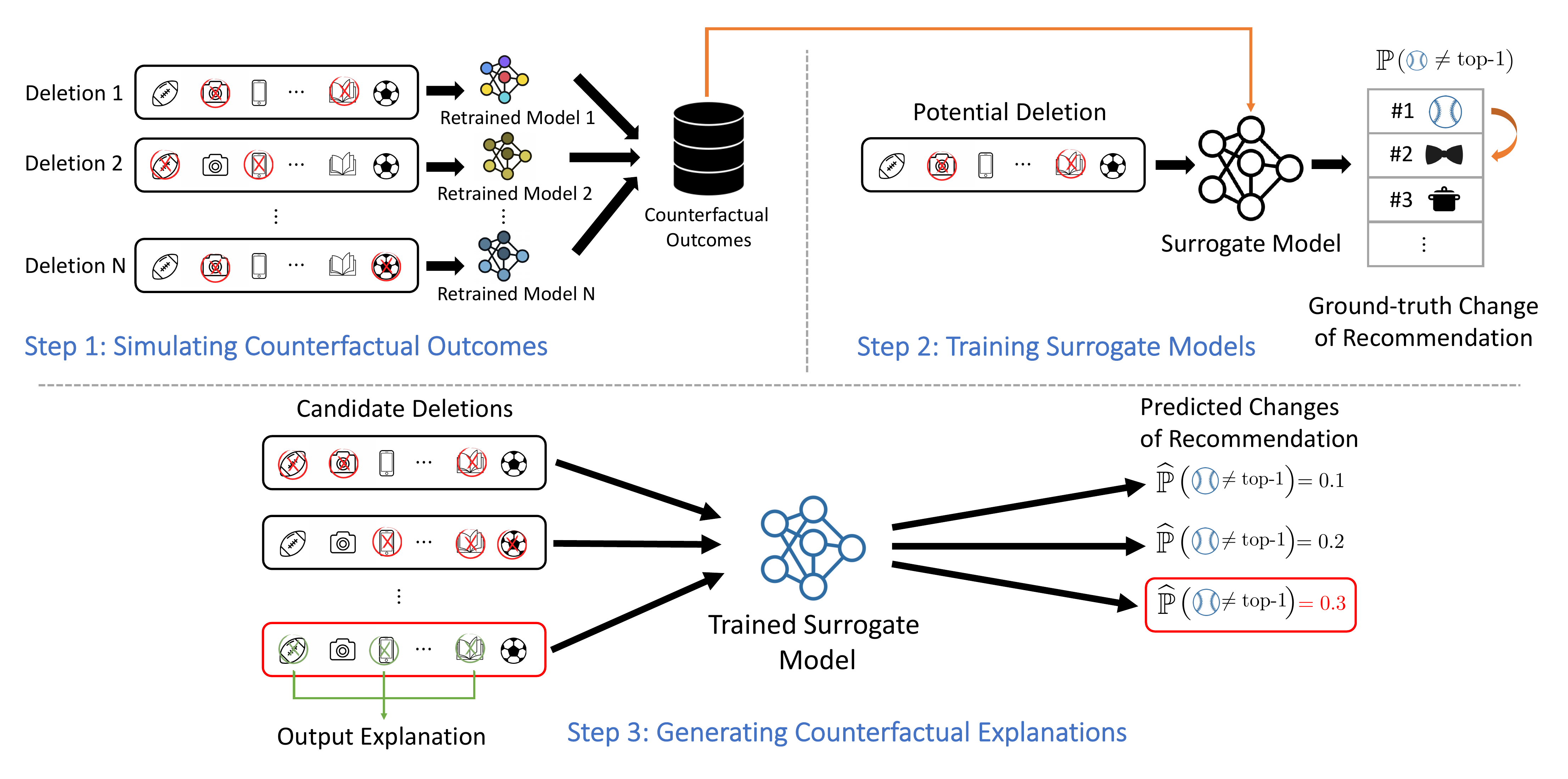}
  \caption{Overview of our proposed method. Step 1: \textbf{Simulating Counterfactual Outcomes}-we empirically simulate the counterfactual outcomes by removing subsets of user history, retraining the model, and observing the change of recommendation. Step 2: \textbf{Training Surrogate Models}-we train a surrogate model to predict how a recommendation would change given a deleted history. Step 3: \textbf{Generating Counterfactual Explanations}-we generate explanations by searching for the deleted history that is predicted by the surrogate model to have the maximum chance to remove the recommendation.
  }
  % \vspace{-0.13cm}
  \label{fig:overview}
\end{figure}

% So far, only a few works study how to generate counterfactual explanations in recommender systems. One approach uses Influence Function~\cite{koh2017understanding} to estimate how the predicted score on the recommendation might change if some training items were removed~\cite{tran2021counterfactual}. However, Influence Function is shown to be relatively inaccurate on deep models~\cite{basu2020influence}, and therefore it is unclear how counterfactually valid the generated explanation can be in practice. We compare with influence function in our experiments.
To this end, we propose a learning-based framework to generate counterfactual explanations. Our key insight is to train a surrogate model to learn the mapping from a change in the recommendation model's training inputs (\ie user history) to its impact on the recommendation. Figure~\ref{fig:overview} shows the overview of our method. We first empirically simulate the counterfactual events repeatedly by removing subsets of user history, retraining the model, and observing the change in recommendation. We then train a surrogate model to predict how a recommendation would change given a deleted history. Once we have the surrogate model, we can replace every expensive actual retraining with a surrogate model inference that estimates the effect of deletion on the recommendation. Finally, we generate explanations by searching for the deleted history that is predicted by the surrogate model to have the maximum chance to remove the current recommended item from the recommendation list. 

Our approach shifts the computational bottleneck from generating explanations on-the-fly to simulating counterfactual outcomes (\ie step 1), which is parallelizable and can be performed offline. In addition, our method reduces the number of retraining needed by relying on the surrogate model's generalizability. We only need to simulate a fraction of all possible deletions that are enough to train the surrogate model, and then rely on the surrogate model to generalize to unseen deletions. Furthermore, our method can be applied to any recommendation model. Through extensive experiments and user studies, we show that our method can generate more counterfactually valid explanations that are considered more satisfactory by users than baselines.

\section{Problem Formulation}
In a recommender system, let $\mathcal{U}$ denote the set of all users (\ie all user IDs),  $\mathcal{I}$ denote the set of all items (\ie all item IDs),  $\mathcal{R}$ denote the set of all ratings (\eg 1-5 stars), and $\mathcal{S} \subseteq \mathcal{U} \times \mathcal{I} \times \mathcal{R}$ denote the history of all user-item interactions with the ratings used as the training set of the recommendation model $\hat{\theta}$.
 
\para{Definition of Counterfactual Explanation.} For a user $u \in \mathcal{U}$ who interacted with a set of items $I_u :=\{i \in \mathcal{I}: (u, i, .) \in \mathcal{S}\} $ in the training set (\ie $u$'s visiting history), if the recommendation model $\hat{\theta}$ recommends an item $i$ to user $u$, we want to find a subset of $u$'s visiting history $E_{u, i} \subseteq I_u $ such that if $E_{u, i}$ were removed from the training set $\mathcal{S}$, then the model $\hat{\theta}$ would not recommend item $i$ to user $u$.

Formally, if we remove an arbitrary subset of user $u$'s history $D_u \subseteq I_u $ and retrain the model on the training set with $D_u$ deleted, the resulting counterfactual recommendation model is\footnote{Technically $\hat{\theta}(.)$ should have two inputs $u$ and $D_u$ to represent the counterfactual model, we fold $u$ into $D_u$ to simplify the notation.}:
\begin{equation}
\label{eq:cf_model}
% \resizebox{0.92\linewidth}{!}{
\hat{\theta}(- D_u) \vcentcolon= \argmin\limits_{\theta \in \Theta} \sum\limits_{(\tilde{u}, \tilde{i}, \tilde{y}) \in S}{ \Big [1 - \mathbb{I}(\tilde{u} = u \text{ and }\tilde{i} \in D_u) \Big ] \cdot \ell(\tilde{y}, f(\tilde{u}, \tilde{i}; \theta))}
% }
\end{equation}

where $\mathbb{I}(.)$ is binary indicator function, $\tilde{y}$ is rating from user $\tilde{u}$ on item $\tilde{i}$ in the training set $\mathcal{S}$,  $f(\tilde{u}, \tilde{i}; \theta)$ is the predicted score on item $\tilde{i}$ for user $\tilde{u}$ by the model $\theta $ , and $\ell(.)$ is recommendation model's training loss.

Suppose the recommender system shows top-$K$ items as the recommendation list to all users, let $\pi_{u, \hat{\theta}}$ denote the ranking list of the items for user $u$ produced by the original recommendation model $\hat{\theta}$ and let $\pi_{u, \hat{\theta}}(i)$ denote the rank of item $i$ in $\pi_{u, \hat{\theta}}$, then an explanation $E_{u, i}$ is defined to be counterfactual if:
\begin{equation}
\label{eq:cf_def}
\pi_{u,  \hat{\theta}(- E_{u, i})}(i) > K
\end{equation}
In other words, the condition is the counterfactual model ranks the recommended item $i$ at a position lower than $K$, which is the minimal rank decided by the recommender system for an item to appear in $u$'s recommendation list.

\para{Naive Approach.} The naive way to generate $E_{u, i}$ is to exhaustively search on all possible subsets of $I_u$, \ie the power set $2^{I_u}$, as shown in Algorithm~\ref{algo:naive} in Appendix~\ref{app:naive}. It is, of course, forbiddingly expensive. To generate an explanation for one user and one received recommended item, the worst case is to retrain the model $O(2^{|I_u|})$ times. If we want to generate explanations for every user's top-$K$ recommendation list, we would need to retrain $O(2^{\overline{|I_u|}} \times |\mathcal{U}| \times K)$ times where $\overline{|I_u|}$ is the average number of items users interacted in the training set.

\para{Model Reuse Trick.} One simple way to reduce the number of retraining times required is to reuse the counterfactually retrained models. Because for a user $u$ and a removed subset $D_u$, the counterfactual model $\hat{\theta}(- D_u)$ is the same for any item $i \in \mathcal{I}$, we can use the same $\hat{\theta}(- D_u)$ to explain all items for the user $u$ if $D_u$ were removed. This reduces the number of retraining times to $O(2^{{\overline{|I_u|}}} \times |\mathcal{U}|)$. We use this trick in our method.

\section{Insights and Overview}
In this section, we explain the potential challenges of generating counterfactual explanations, key insights into our approach, and an overview of our method.

\subsection{Key Insights}
\para{Challenges.} There are three main challenges to generating counterfactual explanations. \textit{First}, how to scale? As illustrated by the naive approach, finding such explanations can be computationally costly. \textit{Second}, how to make the method be applicable to any type of recommendation model? Ideally we do not want to change the existing recommendation module so that the explanation module can be independently deployed and maintained. \textit{Third}, how to describe the counterfactual events? Because machine learning lacks the proper language to model counterfactual events since the majority of ML training algorithms only concern what has happened empirically and factually rather than the ``what-if'' scenarios. One popular approach is causal graphs~\cite{pearl2009causality}. However, there are numerous variables and confounders in today's complex recommender system that makes such causal graph design non-trivial.

\para{Insights.} The key insight is to turn the problem into a learning problem. To generate counterfactual explanations, all we need to know is \textit{how model outputs would change after removing a subset of user history in the training set}. 
% Specifically, we can directly predict the change of model outputs given which part of the training set is removed. 
Specifically, given an event of removing a training subset, we can directly predict its effect on the model outputs.
If we can build a \textit{surrogate model} to predict such effects, then by querying the surrogate model, we can estimate the change of model outputs on the recommended item for a particular user when removing an arbitrary subset of his or her history. Hence our search would be much faster by replacing every expensive model retraining with cheap surrogate model inference. In addition, the whole process speeds up by not having to retrain every time to explain to every user every possible recommended item. Instead, we can only retrain on a subset of all possible deletions to generate enough surrogate model's training data, and then rely on the surrogate model's generalizability to unseen deletions.

\subsection{Methodology Overview}
Our method has three main steps as the following:
\begin{enumerate}
    \item \textbf{Simulating Counterfactual Outcomes}: We empirically obtain \textit{ground-truth} of counterfactual outcomes. We sample a group of training subsets, remove them from the training set, and retrain the recommendation model to get the counterfactual models. We then record counterfactual models' prediction outputs. This step generates the training data for the surrogate model.
    \item \textbf{Training Surrogate Models}: Given the empirically simulated data that describes the mapping between which part of the user history is removed and how the model's predicted output would change consequently, we design and train surrogate models to learn the mapping.
    \item \textbf{Generating Counterfactual Explanations}: Given a trained surrogate model, we generate the counterfactual explanations by searching for the subset of user history predicted by the surrogate model that, if were deleted, would have the maximal chance of removing the recommended item from the user's recommendation list.
\end{enumerate}

\para{Advantages.} \textit{First}, compared to the naive approach, our method requires less retraining by relying on the surrogate model’s generalizability. We only need to repeat retraining a certain number of times that are enough to generate sufficient training data for the surrogate model.
% a subset of all possible deletions as the training set to train the surrogate model,
Then we can rely on surrogate models to generalize to unseen deletions (including unseen users and items). Note that the most computationally expensive step is the first step that simulates counterfactual outcomes, because it still needs to repeatedly retrain the model although the number of retraining times required is much smaller. However this step is perfectly parallelizable and can be performed offline. In other words, we can shift the major computational burden from online to offline. \textit{Second}, our method is applicable to any recommendation models because we only need to know how the recommender model's inputs and outputs change regardless of what the model does within itself, \ie treating the model as a black-box.

% \textit{Third}, we use the mapping from change in user history to change in recommendation to describe the counterfactual events.

\section{Methodology}
In this section, we explain the three steps to generate counterfactual explanations in detail.

\subsection{Step 1: Simulating Counterfactual Outcomes}

% \begin{figure}[!t]
%   \centering
%   \includegraphics[width=0.9\linewidth]{figs/score.pdf}
%   \caption{Definition of our model output change. To quantify how much the target item $i$ would change if a subset of training data were deleted, we look at the score distance, in the counterfactual ranking list, between the item $i$ and the item ranked at item $i$'s original position. In other words, we define the output change as the relative positional change on item $i$ with respect to its original position. This is what the surrogate models predict. \kevin{TODO: add classification.}}
%   \label{fig:score}
% \end{figure}

% \begin{figure}[!t]
%  \centering
%     \subfloat[\centering Regression ]{{\includegraphics[width=0.49\linewidth]{figs/score.pdf} }}%
%     \hfill
%     \subfloat[\centering Classification ]{{\includegraphics[width=0.49\linewidth]{figs/score.pdf} }}%
%     \caption{Two definitions of counterfactual change on an item $i$, \ie what the surrogate model predicts. (a): Regression definition: score change between the item $i$ and the item ranked at item $i$'s original position in the counterfactual ranking list, \ie relative positional change on item $i$ with respect to its original position. (b): .}%
%     \label{fig:score}
% \end{figure}

\begin{figure}[!t]
  \centering
  \includegraphics[width=0.85\linewidth]{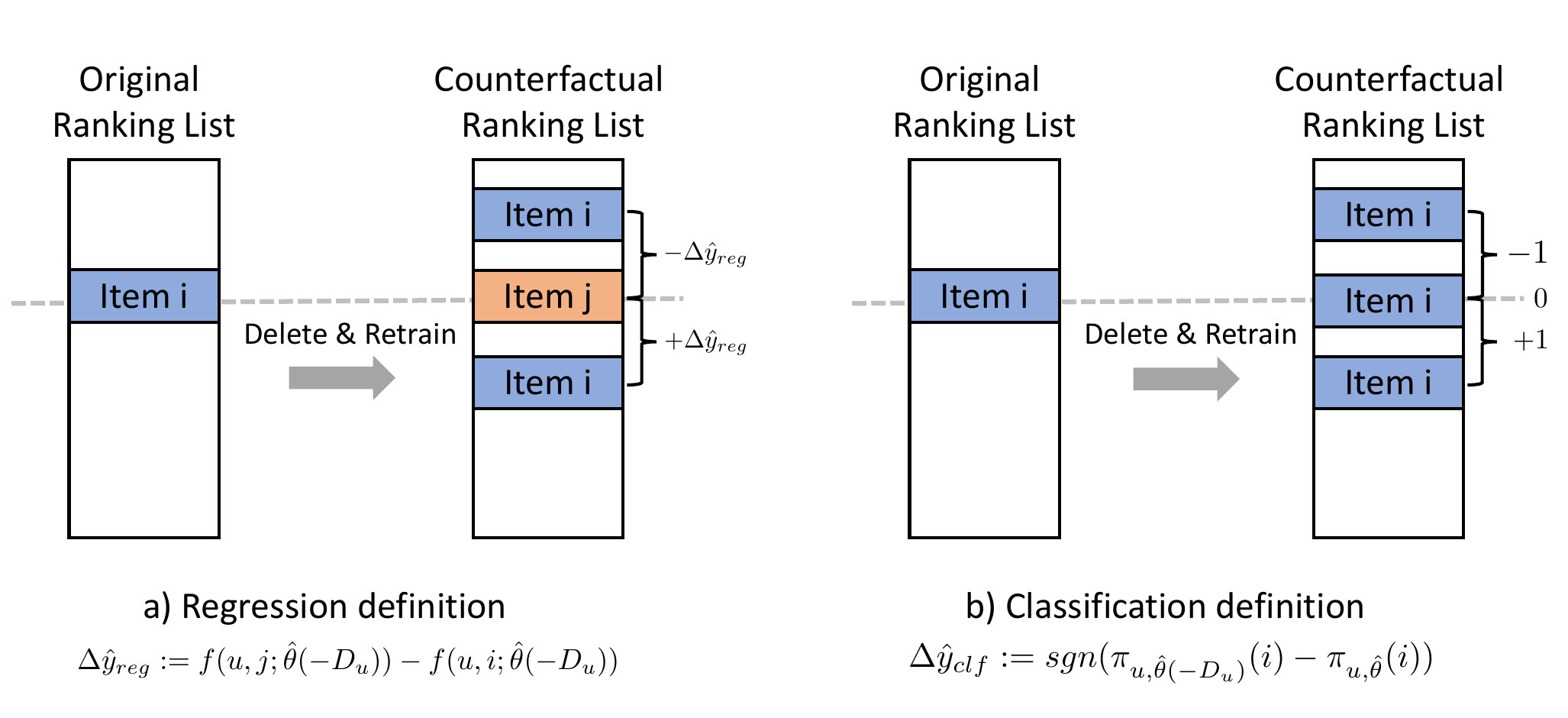}
  \caption{Two definitions of model output change on item $i$ caused by history deletion, \ie the effect that the surrogate models predict. (a): Regression definition: predicted score difference between item $i$ and the item ranked at item $i$'s original position in the counterfactual ranking list (item $j$ in the figure). (b): Classification definition: Direction of rank change (up, down, or unchanged) on item $i$.}
  \label{fig:score}
  % \vspace{-0.4cm}
\end{figure}

% The criteria of counterfactual definition in eq (\ref{eq:cf_def}) is not easily achievable since the item rank is discrete and therefore hard to predict by the surrogate model. Therefore we measure the change of model output by item score, \ie we try to predict how much item $i$'s score would change in the counterfactual model.

We first define the change in model outputs, denoted as $\Delta \hat{y}$. It is what we measure on counterfactual models after retraining to build the dataset that describes the counterfactual outcomes. It is also later what surrogate models predict. Specifically, $\Delta \hat{y}$ describes the following: for a given user $u$, if we were to remove an arbitrary subset $D_u$ from his or her history, how would the model prediction on item $i$ change consequently?

We propose two definitions of model output change on item $i$. The first definition focuses on the difference between predicted scores, and therefore $\Delta \hat{y}$ is a numerical value, and the resulting surrogate model is a regression model. The second definition focuses on the direction of rank change (up, down, or unchanged) which is a discrete class, and therefore the corresponding surrogate model is a classification model.

\para{Regression Definition of $\Delta \hat{y}$.} One straightforward way to define counterfactual change on the target item $i$'s predicted score is to compare the item $i$'s counterfactual score with its original score. 
% However, it might be problematic because the relative magnitude of score would be different between two different ranking lists. 
However the comparison is unfair because two different ranking lists have different score ranges.
Instead we find an item as the benchmark that represents item $i$'s original position in the counterfactual list, \ie the item that takes item $i$'s original position in the counterfactual list, illustrated in Figure~\ref{fig:score} (a).
If item $i$ is ranked at the $k$-th position in the original ranking list, then we look at the item ranked at the $k$-th position in the counterfactual ranking list as the benchmark. And we compute the score difference in the counterfactual ranking list between item $i$ and the benchmark item. If item $i$'s score is lower than the benchmark item's, then it means it is now ranked lower by the counterfactual model than the original model, and therefore item $i$ is more likely to be removed from the recommendation in the counterfactual world.

Formally, recall that $\pi_{u, \hat{\theta}}$ is the original ranking list of the items for user $u$ by the model $\hat{\theta}$; $\pi_{u, \hat{\theta}}(i)$ is the rank of item $i$ in the original ranking list $\pi_{u, \hat{\theta}}$,  $\pi_{u, \hat{\theta}(-D_u)}$ is the counterfactual ranking list generated by the counterfactual model $\hat{\theta}(-D_u)$. Further define $\pi_{u, \hat{\theta}(-D_u)}^{-1}(k)$ to be the item at $k$-th rank in the counterfactual ranking list $\pi_{u, \hat{\theta}(-D_u)}$, the regression definition of model output change on item $i$ for user $u$ after removing $D_u$ is:
\begin{equation}
\label{eq:score_chg}
\begin{aligned} 
& \Delta \hat{y}_{reg}(u, D_u, i) \vcentcolon= f \Big (u, j; \hat{\theta}(-D_u)\Big) - f\Big(u, i; \hat{\theta}(-D_u)\Big), \\ 
& \text{where $j \vcentcolon= \pi_{u, \hat{\theta}(-D_u)}^{-1}(\pi_{u, \hat{\theta}}(i) )$}
\end{aligned} 
\end{equation}

If the predicted score is normalized into $[0, 1]$, then the range of $\Delta \hat{y}_{reg}(.)$ is $[-1, 1]$. In addition, if $\Delta \hat{y}_{reg}$ is positive, it means item $i$ is now ranked lower than its original position after removing $D_u$; if it is negative, it means higher; if zero, it means the rank does not change. Therefore, a larger $\Delta \hat{y}_{reg}$ means, in the counterfactual case, the model is less likely to recommend $i$ for $u$, and therefore item $i$ is more likely to disappear from $u$'s recommendation list after removing $D_u$. Hence we want to maximize $\Delta \hat{y}_{reg}$ when generating counterfactual explanations. 

\para{Classification Definition of $\Delta \hat{y}$.} An alternative definition is to look at the direction of rank change (up, down, or unchanged) in item $i$. In this case, we only compare item $i$'s position with itself between the original and counterfactual ranking list, shown in Figure~\ref{fig:score}(b). The resulting change is a qualitative class rather than a quantitative score:

\begin{equation}
\label{eq:score_clf}
\Delta \hat{y}_{clf}(u, D_u, i) \vcentcolon= sgn\big(\pi_{u, \hat{\theta}(-D_u)}(i) - \pi_{u, \hat{\theta}}(i)\big)
\end{equation}

Note that there might be some alternative ways to define model output change, see Section~\ref{subsec:abla} for ablation studies.

\para{Generating Surrogate Model's Training Set.} We describe how to generate surrogate model's training set, summarized in Algorithm~\ref{algo:generate}. We first sample a subset from all training users. Then for each user $u$, we sample a group of subsets from user $u$'s history. For each history subset, we remove it from the full training dataset and retrain the recommendation model to get the counterfactual model. After it, we generate the counterfactual ranking list for user $u$ based on the counterfactual model. Given the counterfactual list, we need to decide which item's change $\Delta \hat{y}$ to include in the surrogate model's training set. Technically, we can include all items in the system (recall the model reusing trick, for the same user, we do not need to retrain a different model to get the score change on a new item) but it is unnecessary because eventually we only need to explain items recommended to the user. Therefore we only need to include items likely to be recommended, \ie top items. Assume top-$N$ items are shown to the user, then we only need to compute the score change on them as the training set.

\begin{algorithm}[t]
\begin{algorithmic}[1]
\State \textbf{Input:} Recommendation model $\hat{\theta}$ and its training dataset $\mathcal{S}$, rank threshold $N$ that determines recommendation.
\State \textbf{Output:} Counterfactual training dataset $\mathcal{D}_{cf}$.
\State $\mathcal{D}_{cf} \leftarrow \emptyset$.
\State Sample a subset of users $\mathcal{U}^{tr}$ from $\mathcal{U}$ in $\mathcal{S}$.
\For{$u \in \mathcal{U}^{tr}$}
	\State Sample a group of history subsets $\mathcal{D}_u^{tr}$ from $u$'s history $I_u$.
	\For{$D_u \in \mathcal{D}_u^{tr}$}
	    \State Retrain to obtain $\hat{\theta}(-D_u)$ with eq(\ref{eq:cf_model}).
	    \For{$i \in \pi_{u, \hat{\theta}}[:N]$}  //  Top-$N$ items by $\hat{\theta}$.
	       % \State Compute $\Delta \hat{y}(u, D_u, i)$ with eq(\ref{eq:score_chg}) or eq(\ref{eq:score_clf}).
	        \State $x_{cf} \leftarrow (u, D_u,  i)$.
	        \State $y_{cf} \leftarrow \Delta \hat{y}(u, D_u, i)$. // By eq(\ref{eq:score_chg}) or (\ref{eq:score_clf}) with $\hat{\theta}(-D_u)$.
	        \State $\mathcal{D}_{cf} \leftarrow \mathcal{D}_{cf} \cup \{(x_{cf}, y_{cf})\}$.
	    \EndFor
	\EndFor
\EndFor
\State Return $\mathcal{D}_{cf}$. 
\end{algorithmic}
\caption{Counterfactual Data Generation.}
\label{algo:generate}
\end{algorithm}

\subsection{Step 2: Training Surrogate Models}
$\Delta \hat{y}$ is a complex function that involves \textit{training and evaluating the recommendation model on different training subsets}, and therefore cannot be directly formulated and obtained. The goal of the surrogate model is to empirically approximate $\Delta \hat{y}$ in an end-to-end manner. 
% \ie we want to predict the change on item $i$ for user $u$ when a history subset $D_u$ is removed. 
The input space of the surrogate model is large if we consider learning the effect on every item caused by every user's every possible deletion of his or her interactions. Recall $\mathcal{S}$ is the interactions in the training set, the entire input space of the surrogate model would be $2^{|\mathcal{S}|} \times |\mathcal{I}| \times |\mathcal{U}|$. And therefore it would be difficult to generate enough training instances to learn the mapping well.

% Recall $\mathcal{S}$ is all interactions in the system. The map from removal set to score change set is:
% \begin{equation}
% \mathcal{U} \times 2^\mathcal{S}  \times \mathcal{I}  \rightarrow \mathbb{R}
% \end{equation}

\para{User Independence Assumption or SUTVA.} Recall that when we define the counterfactual explanation (Figure~\ref{fig:exp_def}), we implicitly assume that the user $u$'s recommendation is only impacted by $u$'s own history rather than other users' history. This is also known as \textit{Stable Unit Treatment Value
Assumption} in the causal inference literature~\cite{angrist1996identification}. Technically speaking, a user's recommendation is also dependent on other users' histories. But it would be too complex to serve as an explanation. Consider the oddity of the following explanation: ``Had some other strangers that you do not know interacted with those items, then we would not recommend it.'' 
% or ``Had this item not been visited by some strangers you do not know, then we would not recommend it.'' 
Note that regarding this assumption, we merely follow the conventional item-based collaborative-filtering explanation, which also assumes a user's recommendations only depend on his or her own history (``We recommend this item because \textit{you (and only you)} have visited the following items.'').

Given this assumption, we can reduce the sampling space of deletion from $2^{|\mathcal{S}|}$ to $2^{\overline{|I_u|}}$ where $\overline{|I_u|}$ is the average number of items users interacted in the training set. Therefore the form of regression surrogate model is:
\begin{equation}
\alpha_{reg}: \mathcal{U} \times 2^{\overline{{|I_u|}}}  \times \mathcal{I}  \rightarrow \mathbb{R}
\end{equation}

And the form of classification surrogate model is:
\begin{equation}
\alpha_{clf}: \mathcal{U} \times 2^{\overline{{|I_u|}}}  \times \mathcal{I}  \rightarrow \{-1, 0, 1\}
\end{equation}

We formalize the training process of the surrogate model that empirically approximates $\Delta \hat{y}$. Let $\mathcal{D}_{cf}$ be the training data collected in Algorithm~\ref{algo:generate} from the previous step, \ie $\mathcal{X}_{cf} := \{x_{tr}: (x_{tr} , \cdot) \in \mathcal{D}_{cf}  \}$ be the set of training inputs (collected in Line 10 Algorithm~\ref{algo:generate}), $\mathcal{U}^{tr} := \{u \in \mathcal{U}: (u, \cdot, \cdot) \in \mathcal{X}_{cf}\}$ be the training users collected, $\mathcal{D}_u^{tr} := \{D_u \in 2^{I_u}: (u, D_u,\cdot) \in \mathcal{X}_{cf}\}$ be the sampled deleted subsets of user $u$'s history and $\mathcal{I}^{tr}_{u, D_u} := \{i \in \mathcal{I}: (u, D_u,i) \in \mathcal{X}_{cf}\}$ be top-N items whose $\Delta \hat{y}$ we collect for user $u$'s deletion $D_u$. Then the surrogate model can be trained with the following Empirical Risk Minimization:
\begin{equation}
% \resizebox{0.9\linewidth}{!}{%
\hat{\alpha} \vcentcolon= \argmin\limits_{\alpha \in \Alpha}{\sum\limits_{u \in \mathcal{U}^{tr}}}\sum\limits_{D_u \in \mathcal{D}_u^{tr} }\sum\limits_{i \in \mathcal{I}_{u, D_u}^{tr}}{\ell_{\alpha} \big (g(u, D_u, i; \alpha),  \Delta \hat{y}(u,D_u, i) \big )}
% }
\end{equation}
where $g(.;\alpha)$ is the surrogate model $\alpha$'s prediction; $\ell_{\alpha}(.)$ is the loss of the surrogate model, which is MSE for the regression and cross-entropy for the classification; $\Delta \hat{y}(u, D_u, i)$ is the corresponding ground-truth value of $\Delta \hat{y}$ collected.

\para{Input Representation.} An important design choice is how to represent the surrogate model's three inputs $u$ (target user), $D_u$ ($u$'s deleted history), and $i$ (target item). To represent $u$ and $i$, we simply use the recommendation model's generated embeddings (or any user/item embeddings that exist in the recommender system). It is a common practice in recommender system because the learned embeddings usually can represent user and item information well. In terms of deleted items ($D_u$) which is an item set, we use a simple heuristic: the sum of the embedding of the deleted items. Formally, let $\phi_u(u) \in \mathbb{R}^p$ be the user embedding of user $u$, and $\phi_i(i) \in \mathbb{R}^q$ be the item embedding of item $i$, we simply concatenate those three features into a single feature vector as the surrogate model's training input as the following:
\begin{equation}
\big [\phi_u(u) \quad  \sum_{\tilde{i} \in D_u}\phi_i(\tilde{i}) \quad \phi_i(i) \big ]
\end{equation}

We experiment with different representations of $D_u$, for example DeepSets~\cite{zaheer2017deep} (one can view this sum of embeddings as DeepSets with identity mapping), and find this simple method works well. See Section~\ref{subsec:abla} for the related ablation studies.

\para{Choosing Surrogate Models.} Technically speaking, the surrogate model can be any regression or classification model. We choose LASSO~\cite{tibshirani2011regression} and a simple 3-layer MLP for regression and logistic regression and the same MLP architecture for classification. We empirically find that those simple models outperform deep and complex models (in terms of the generated explanation's ability to satisfy the counterfactual definition). See Section~\ref{subsec:abla} for related ablation studies.

% \begin{table*}[!t]
% \resizebox{0.9\linewidth}{!}{
% \begin{tabular}{|l|llll|l|lll|}
% \hline
% \multirow{2}{*}{Experiment} & \multicolumn{4}{l|}{Dataset} & \multirow{2}{*}{Recommender Model} & \multicolumn{3}{l|}{Model Performance} \\ \cline{2-5} \cline{7-9} 
%  & \multicolumn{1}{l|}{Name} & \multicolumn{1}{l|}{\# of Users} & \multicolumn{1}{l|}{\# of Items} & \# of Interactions &  & \multicolumn{1}{l|}{NDCG@10} & \multicolumn{1}{l|}{Precision@10} & Recall@10 \\ \hline
% \textit{MovieLens} & \multicolumn{1}{l|}{MovieLens-100K~\cite{mvledata}} & \multicolumn{1}{l|}{943} & \multicolumn{1}{l|}{1682} & 100,000 & Matrix Factorization (SVD)~\cite{} & \multicolumn{1}{l|}{0.0959} & \multicolumn{1}{l|}{0.0911} & 0.0327 \\ \hline
% \textit{Netflix} & \multicolumn{1}{l|}{Netflix (Small)~\cite{netflix}} & \multicolumn{1}{l|}{10,000} & \multicolumn{1}{l|}{5,000} & 607,803 & Bayesian Personalized Ranking~\cite{} & \multicolumn{1}{l|}{0.2815} & \multicolumn{1}{l|}{0.2201} & 0.1846 \\ \hline
% \textit{Amazon} & \multicolumn{1}{l|}{Amazon Digital Music~\cite{netflix}} & \multicolumn{1}{l|}{5,541} & \multicolumn{1}{l|}{3,568} & 64,706 & Neural Collaborative Filtering~\cite{} & \multicolumn{1}{l|}{0.0244} & \multicolumn{1}{l|}{0.0104} & 0.0406 \\ \hline
% \end{tabular}
% }
% \caption{Datasets, recommendation models and the corresponding model performance in our experiments.}
% \label{tab:exp}
% \end{table*}

\subsection{Step 3: Generating Counterfactual Explanations}
\label{subsec:gen}
\para{Fixed Explanation Size.} Conventionally in the counterfactual explanation literature, the explanation size should be as small as possible. However, we argue that, in the context of recommender system, this is not only unnecessary but might even be harmful. Instead, it would be better to have fixed-size explanations. It is because displaying an uncontrollable number of items when explaining to users would mess up the user interface and make users feel overwhelmed and incomprehensible if showing too many items. In fact, a fixed explanation size is the common design choice of the majority of \textit{item-based collaborative filtering} explaining logic~\cite{resnick1994grouplens,herlocker2000explaining}.

Given a fixed explanation size $e$, the trained surrogate model $\hat{\alpha}$, target user $u$ and target item $i$, to generate an explanation we first randomly sample a fraction of all possible $e$-size subsets in $u$'s history $I_u$, and then search for the history subset with maximal \footnote{Recall that in the regression definition, larger predicted $\Delta \hat{y}$ means target item $i$ is more likely to be ranked lower than the item ranked at its original position, and therefore more likely to satisfy the counterfactual definition. In the classification definition, larger predicted $\Delta \hat{y}$ (which is the classification surrogate model's logit) means target item $i$'s rank is more likely to be larger than its original rank, and therefore is more likely to disappear from the recommendation.} predicted $\Delta \hat{y}$ as the generated explanation, \ie
\begin{equation}
\label{eq:gen}
E_{u, i}^* = \argmax_{E \sim 2^{I_u}; |E| = e}{g(u, E, i ; \hat{\alpha})}
\end{equation}

Ideally one can improve from this simple subset searching by optimization and approximation techniques. In practice, we find this step is fast enough (See Section~\ref{subsec:results}) because our surrogate model is simple and therefore fast in making inferences. In addition, thanks to the fixed explanation size design, the sampling space is much smaller compared to finding the minimal size . We experiment with some approximation techniques, and find they not only do not significantly speed up the process but also lead to worse performance on the generated explanations. In practice, if the sampling size is too large, we can cap it in the experiments. See Section~\ref{subsec:setup} for details.

\section{Experiments}
\label{sec:eval}
In this section, we show the experimental results that evaluate the proposed method.

\subsection{Setup}
\label{subsec:setup}
\para{Dataset and Recommendation Model.} We use three datasets summarized in Table~\ref{tab:exp} in Appendix~\ref{app:data}. For each dataset, we experiment with two recommendation models: Matrix Factorization (SVD) and Neural Collaborative Filtering~\cite{he2017neural}. We randomly split all datasets into $70\%$ training and $30\%$ test set.

% and train a different type of recommender model on each dataset. We randomly split all datasets into $70\%$ training and $30\%$ test set.  summarizes the datasets, models, and the corresponding test performance. We name those three experiments as \textit{MovieLens}, \textit{Netflix}, and \textit{Amazon}.

\para{Explanation Generation.} In \textit{Simulating Counterfactual Outcomes} step, we randomly sample $200$, $100$, and $80$ users from \textit{MovieLens}, \textit{Netflix}, and \textit{Amazon} respectively as surrogate model's training users. For each training user $u$, we randomly remove $1,000$ 3-item subsets from $u$'s history (all deleted history subsets have the same size) and retrain the model to obtain counterfactual models. The resulting number of retrained counterfactual models is $200K$, $100K$, and $80K$ for \textit{MovieLens}, \textit{Netflix}, and \textit{Amazon} respectively. For each counterfactual model, we generate $\Delta \hat{y}$ on the top-$20$ (\ie $N=20$ in Algorithm~\ref{algo:generate}) recommended items for the target user. In \textit{Training Surrogate Model} step, when we train LASSO for the regression model and logistic regression for the classification model, we use k-folder cross validation to tune the hyper-parameters. In \textit{Generating Counterfactual Explanations} step, we choose the explanation size to be the same as the size of each deleted history subset in \textit{Simulating Counterfactual Outcomes} stage, which is 3 items (\ie $e = 3$ in eq (\ref{eq:gen}), see Section~\ref{subsec:abla} for related ablation study), and we randomly sample at most $100K$ explanation candidates to search for the one with maximum surrogate model prediction.

\para{Evaluation Metric.} As mentioned in Section~\ref{subsec:gen}, we argue fixed-size explanations might be more beneficial from the perspective of implementation and user comprehension. Therefore aiming for a small explanation size is out of our evaluation scope, and we focus on the counterfactual validity, which means if the explanations were indeed removed from the training set, the user actually would not receive the recommendation. Therefore we judge a counterfactual explanation is valid if it removes the recommended item from the user's recommendation list, \ie eq (\ref{eq:cf_def}). How many items are shown to users in a recommendation list is decided by the design of the recommender system. For example, if a recommender system shows the top-$K$ items to users, then an explanation is valid if the current item would fall out of the top-$K$ ranking list. We evaluate different choices of $K$.

In all experiments, we sample $200$ test users disjointly from the training users included in the surrogate model's training set. For each user $u$, we choose the target item that we explain as $u$'s top-1 item. Then we remove the generated explanation w.r.t to the target item from the user's history $I_u$, and retrain the model on the deleted training set to obtain the ground-truth counterfactual model for evaluation. After it, we look at the ground-truth counterfactual model's top-$K$ recommendation on the target user to see if it still includes the target item or not. All the explanations generated by our method or baselines have size 3 items.

\para{Baseline.} We consider two baselines. The first is \textit{k-nearest neighbors} (KNN) which searches for the items in the history that are the nearest neighbors of the target item in the item embedding space. The second is \textit{Influence Function}, which is a generic explaining approach that can be used to estimate the impact of removing a training point on a test point without fully retraining the model. The influence function in~\cite{koh2017understanding} is defined on the loss of the test point, which is, in our case, the target item $i$. However its ground-truth rating is unavailable, and therefore we cannot compute the model's loss on it. Similar to ~\cite{fia,tran2021counterfactual},
% (eq (12) in~\cite{fia}), 
we adapt the influence function's definition for the recommendation by defining it on the test point's predicted score. For the target user $u$ and the target item $i$, the impact of removing any training item $\tilde{i}$ is:
    
\begin{equation}
Infl(u, i, \tilde{i}) := \frac{1}{n}\nabla_\theta f(u, i; \hat{\theta})^{\intercal}H_{\hat{\theta}}^{-1}\nabla_{\theta} L(u, \tilde{i}; \hat{\theta})
\end{equation}

where $n$ is the number of training samples, $H_{\hat{\theta}}$ is Hessian of the original model $\hat{\theta}$ and $L(u, \tilde{i}; \hat{\theta})$ is the loss on the training item $\tilde{i}$. We compute the influence function on all $\tilde{i} \in I_u$ and select items with maximal influence w.r.t the target item $i$ (\ie removing them would decrease the predicted score on $i$ the most, as predicted by the Influence Function) as the generated explanation.
% \begin{itemize}
%     % \item \textbf{Random}: We randomly sample 3 items from the target user's history.
%     \item \textbf{KNN}: We search for the items in the history that are 3-nearest neighbours of the target item in the item embedding space.
%     \item 
% \end{itemize}

% \begin{table}[!t]
% \resizebox{0.7\linewidth}{!}{
% % \begin{tabular}{|l|l|l|l|l|l|}
% % \hline
% %  & Random & KNN & \begin{tabular}[c]{@{}l@{}}Influence \\ Function\end{tabular} & \begin{tabular}[c]{@{}l@{}}Linear\\ Surrogate\end{tabular} & \begin{tabular}[c]{@{}l@{}}MLP\\ Surrogate\end{tabular} \\ \hline
% % Avg. time (sec) & $6.91 \times 1e{-4}$ & $1.25 \times 1e^{-3}$ & 6.47 & 1.75 & 2.60 \\ \hline
% % \end{tabular}
% \begin{tabular}{@{}lllll@{}}
% \toprule
%  & KNN & Influence Function & Linear Surrogate (reg) & MLP Surrogate (reg) \\ \midrule
% Time (sec) &  $1.25 \times 1e^{-3}$ & 6.47 & 1.75 & 2.60 \\ \bottomrule
% \end{tabular}
% }
% \caption{Average generation time (seconds) per explanation on \textit{MovieLens} with Matrix Factorization model.}
% \label{tab:time}
% % \vspace{-0.5cm}
% \end{table}

\begin{figure}[!t]
\centering
\begin{minipage}{.49\textwidth}
  \centering
  \includegraphics[width=0.9\linewidth]{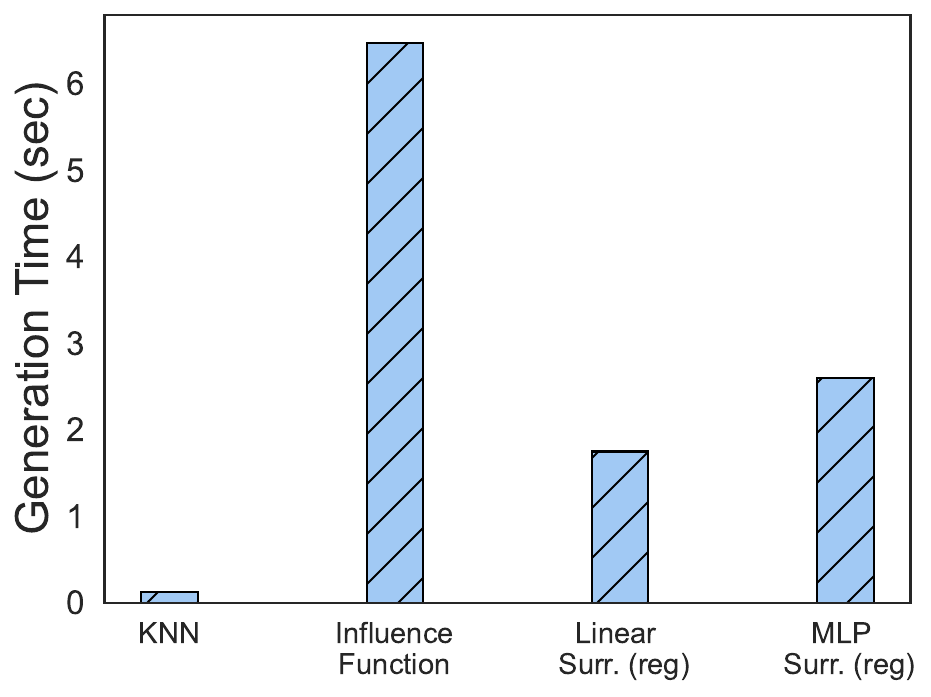}
  \captionof{figure}{Average generation time (seconds) per explanation on \textit{MovieLens} with Matrix Factorization model. Our method (Linear Surr. and MLP Surr.) generate explanations faster than Influence Function but slower than the simple KNN.}
  \label{fig:gen_time}
\end{minipage}%
\hfill
\begin{minipage}{.49\textwidth}
  \centering
  \includegraphics[width=0.95\linewidth]{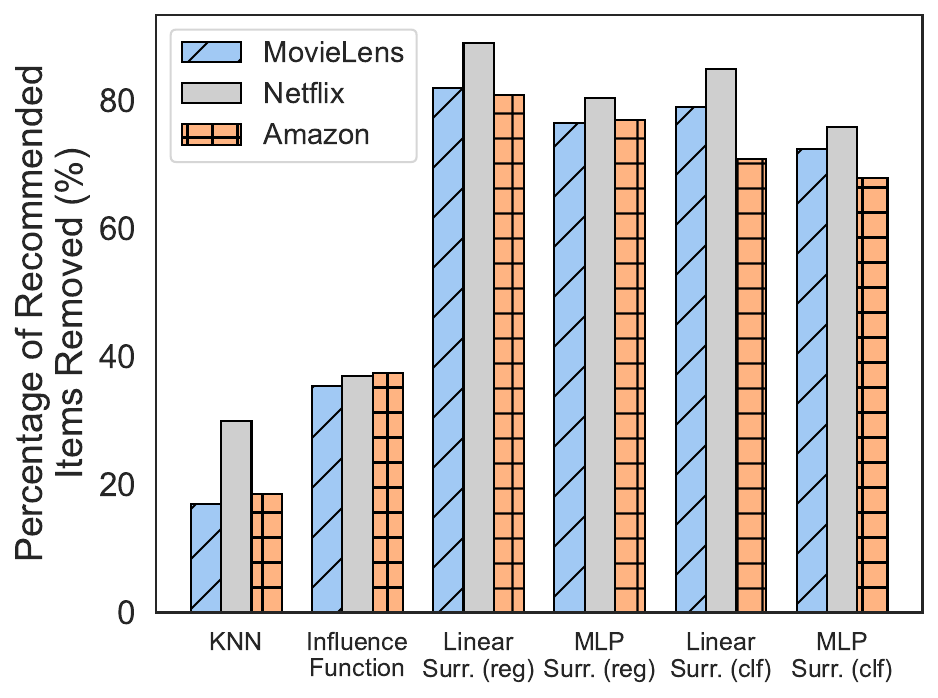}
  \vspace{0.1cm}
  \captionof{figure}{Percentage of (top-1) recommended items that are actually removed from the original position if the explanatory items are indeed removed and the model is retrained. The model is Matrix Factorization trained on different datasets.}
  \label{fig:mf_result}
\end{minipage}
% \vspace{-1.4cm}
\end{figure}

\subsection{Main Results}
\label{subsec:results}

\para{Generation Time.} We report the generation time of explanations in Figure~\ref{fig:gen_time}. Our method generates explanations faster than Influence Function (which is known to be slow in recommendation~\cite{fia}) but slower than the simple KNN. In addition, using MLP as the surrogate model does not significantly slow down the generation because our MLP model architecture is simple (only 3 layers).

\para{Results.} Figure~\ref{fig:mf_result} shows performance of our generated explanations (both regression and classification surrogate model) along with baselines on Matrix Factorization model. We test if the top-1 recommended items are actually removed. See Appendix~\ref{app:full_exp} for full results with different top-K (K=1,3,5) and the results on Neural Collaborative Filtering model. The first observation is our surrogate methods, either regression or classification or linear model or MLP model, outperform the baselines (we will show more analysis shortly after). The second observation is the best performing method is the regression surrogate model. It has two implications. \textit{First}, regression models outperform classification models. This is likely because $\Delta \hat{y}_{reg}$ is a numerical score that can preserve more fine-grained information about how the target item is impacted than the categorical $\Delta \hat{y}_{clf}$. \textit{Second}, linear models outperform MLP models, in both regression and classification, which is consistent with the findings in~\cite{ilyas2022datamodels} that also point out linear surrogate models can predict a complex model's predictions surprisingly well. See Section~\ref{subsec:abla} for an ablation study on linear surrogate models. 

\para{Analysis of Baselines.} We explain why the baselines do not perform well. What all methods aim to approximate ultimately is the decrease of target item $i$'s score if the explanation were removed. An ideal distance metric should be positively correlated with the $i$'s score decrease in order to serve as a valid counterfactual explanation. To validate to how much degree this is the case in KNN and influence function, we randomly sample $500$ item sets from a user's history (on MovieLens dataset with Matrix Factorization model) as explanation candidates, and then measure the average item embedding distance and influence function distance between explanations and the recommended item, and compare them to the ground-truth score decrease of the recommended item after actually removing explanations and retraining the model. In Figure~\ref{fig:dist} (Appendix~\ref{app:baseline}), we can see neither metric is highly correlated with the ground-truth score decrease (with Pearson correlation $0.175$ and $0.356$ only), meaning they cannot serve as good proxies of measuring counterfactual impact on the recommended item, and therefore do not lead to valid counterfactual explanations. On the other hand, since our surrogate model is trained on the \textit{ground-truth} score change obtained by \textit{actual} deletions and retraining, it serves as a more accurate counterfactual measure. This shows obtaining exact ground-truth change on model outputs through retraining is vital to counterfactual validity, even though it is costly.

\subsection{Ablation Studies}
\label{subsec:abla}
To better understand some of our design choices, we show a series of ablation studies. All the following experiments are conducted on MovieLens dataset with Matrix Factorization model.

\para{Definition of $\Delta \hat{y}$.} There could be some alternative ways to define the model output change other than the definition used in eq(\ref{eq:score_chg}) and (\ref{eq:score_clf}). One alternative is the rank difference on the target item $i$ between the original and the counterfactual list (which leads to a regression model that predicts an integer value). See Appendix~\ref{app:score} for the full definition. We find this definition leads to a slightly lower performance than the regression linear surrogate model-the performance when K=1 would decrease by $1.5\%$. This is because the rank difference is an integer, which is less fine-grained than the float score in $\Delta \hat{y}_{reg}$.

Another alternative definition is the score difference of item $i$ between the original and the counterfactual ranking list. Intuitively, if the score becomes lower in the counterfactual list, then one might think it signifies the satisfaction of the counterfactual definition. However, the difficulty of this definition lies in the fact that the original and the counterfactual ranking list have a different score range, and therefore the score difference might not represent the positional difference. One solution is to normalize the score of item $i$ in two lists by the score of the top-1 item in two lists respectively (See Appendix~\ref{app:score} for the complete definition). Our ablation study shows this definition leads to a noticeably worse performance: $7.4\%$ lower than the regression linear surrogate model's performance in \textit{MovieLens} (K = 1). This is because merely normalizing scores does not compensate enough for the score range discrepancy between different ranking lists.

\para{Deletion Size.} In \textit{Simulating Counterfactual Outcomes} step, we need to decide how many items in a history subset to delete when empirically simulating the effect of removal. We choose the removal size to be the same as the fixed explanation size in the final generation step. We experiment with the following alternative: if we remove $40\%$ of a user's history instead of a fixed 3 items, the generated 3-item explanations show a slightly lower performance: $4.5\%$ decrease on the regression linear surrogate model \textit{MovieLens} (K = 1). The number of items deleted in the generation stage has a certain impact on the performance though the impact is not significant.

\para{Representation of $D_u$.} We experimented with more complex set represents of $D_u$, \ie DeepSets~\cite{zaheer2017deep}, and find it leads to $11.5\%$ decrease in the final performance in the regression and linear surrogate model (K = 1). This might suggest that the simple design of input representation is better probably because the surrogate model design is simple.

\para{Analysis of Linear Surrogate Models.} In addition to linear and MLP surrogate models, we experimented with deeper models and more complex input representation, \eg transformer, attention, mask representation of deleted items \etc But eventually we find the simple surrogate models work the best. One observation from Table~\ref{tab:mf_full} in Appendix~\ref{app:full_exp} is the linear models outperform MLP models by a larger margin in Matrix Factorization (5.5--11\% in regression when $K = 1$) than in Neural Collaborative Filtering (0\%--3\%). One hypothesis is Matrix Factorization is a (bi)linear model, and therefore its input-output relationship might be more linear, and therefore easier for the linear surrogate model to learn.
% Note that the linear surrogate model outperforms MLP by a larger margin ($18\%$) in \textit{MovieLens} than the other two experiments. To understand if this supreme performance is caused by the recommendation dataset or the recommendation model, we train a Matrix Factorization recommendation model on Netflix dataset, and then test the performance of the generated explanations under this setting. As we can see in Table~\ref{tab:main}, when we generate explanation on BPR model, the performance of linear surrogate model is comparable to MLP (both are $58\%$ when $K = 1$). But if we replace the BPR with Matrix Factorization, a bi-linear model, then linear surrogate model outperforms MLP with a much larger margin ($32\%$). This might suggest the supreme performance of linear model is more related to recommendation model rather than the dataset, especially when the recommendation model is also (bi)linear, linear surrogate model might learn the mapping more easily.
% One hypothesis is the mapping surrogate model learns tends to be more linear when the recommendation model is linear. 
To verify this hypothesis, we visualize the surrogate model's training data (on MovieLens with Matrix Factorization using $\Delta \hat{y}_{reg}$ definition) via t-SNE~\cite{van2008visualizing} (Figure~\ref{fig:vis} in Appendix~\ref{app:vis}). However we do not observe a clear linear separability. It remains unclear why linear models can predict a complex model's predictions surprisingly well, a phenomenon also pointed out by~\cite{ilyas2022datamodels} as an open problem.

% \kevin{TODO: influence function, generation time, user study}
\begin{table}[!t]
\centering
\resizebox{0.8\linewidth}{!}{
\begin{tabular}{@{}lllll@{}}
\toprule
 & KNN & Influence Function & Linear Surrogate (reg) & MLP Surrogate (reg) \\ \midrule
Avg. Rating & 3.586 $\pm$ 0.74 & 3.658 $\pm$ 0.67 & \textbf{3.803 $\pm$ 0.87} & 3.789 $\pm$ 0.89 \\ \bottomrule
\end{tabular}
}
\caption{Average rating on explanations from the user study. Higher ratings mean more satisfaction. One-sided paired t-test shows explanations generated by the linear surrogate method have higher ratings than KNN, Influence Function, and the MLP surrogate with p-value $0.011$, $0.034$, and $0.393$ respectively.}
\vspace{-0.5cm}
\label{tab:user}
\end{table}

\section{User Study}
\label{sec:user}
To examine the quality of our explanations perceived by the users, we carry out a survey-based user study.

\para{Survey Design.} Our survey contains 5 questions. In each question, we first show participants a list of movies that a user from MovieLens dataset had watched in the past, and then ask participants to imagine being that user. Next we present the top-1 recommended movie by the Matrix Factorization model. Afterward, we show, in random order, four different explanations: KNN, Influence Function, Linear Surrogate (regression), and MLP surrogate (regression). For KNN and Influence Function, the explanation text is ``We recommend this movie because you watched the following movies.'' For linear and MLP surrogate explanations, the explanation text is ``Have you not watched the following movies, we would not recommend this movie.'' Finally we ask participants to rate their satisfaction on each explanation (``Given a scale from 1 to 5, how much are you satisfied with this explanation?'').

\para{Quality Control.} We implement a series of mechanisms to ensure response quality. \textit{First}, we only choose MTurk workers with at least $80\%$ approval rate in participating history (HIT rate) and with at least 50 tasks approved in the past. \textit{Second}, we declare the requirement of movie familiarity to the participants in the advertisement. \textit{Third}, we insert a trivial task (``Please choose both A and D.'') in the middle of the survey, and filter out any responses that fail to answer correctly. \textit{Fourth},  we participants how many movies they usually watch in a month, and remove any participants who answer less than one movie per month. \textit{Lastly}, we ask participants, at the end of the survey, to disclose their familiarity with the movies that they have seen in the survey (``What percentage of movies shown in this survey are familiar to you?''), and exclude participants who claim more than $40\%$ of movies they see are unfamiliar.

\para{Collection.} For $100$ responses we receive, we exclude $8$ users who fail to pass the quality control tests. Of the remaining $92$ participants, $60.9\%$ indicated male and $39.1\%$ indicated female. The majority of participant ages fall into the ranges of 21 to 30 ($38.0\%$) or 31 to 40 ($33.7\%$), with $15.2\%$ falling between 41 and 50, and the rest being older than 50.

\para{Results.} We collect survey results from $100$ users on \textit{Amazon Mechanical Turk}.
% See Appendix~\ref{app:user} for details about our quality control and result collecting. 
Table~\ref{tab:user} shows the average rating on four types of explanations over all questions. \textit{First}, explanations generated by the linear surrogate (regression) model have the highest ratings. \textit{Second}, one-sided paired t-tests show explanation ratings on the linear surrogate method are higher than KNN and Influence Function with statistical significance (with p-value $0.011$ and $0.034$ respectively), but no significance when compared to the MLP surrogate's explanations (with p-value $0.393$).

% \section{Discussion}
% \label{sec:limit}
% We point out two open problems in our work. \textit{First}, do we need to retrain models from scratch in \textit{Counterfactual Data Generation} stage? Can we accelerate this step by using techniques of fast retraining in the area of machine unlearning? \textit{Second}, as mentioned, it is unclear why linear surrogate model generalizes the best in our task. 

% \textit{Third}, as also pointed out by~\cite{ilyas2022datamodels}, we lack the theoretical understanding on why we can predict, to some extent, the model's prediction when its training inputs change.
\section{Related Work}
\label{sec:related}
\para{Explainable Recommender System.} One of the most popular explanation approaches is \textit{collaborative filtering based logic}, which relies on relevant users or items to explain recommendations. In \textit{user-based collaborative filtering}~\cite{resnick1994grouplens,herlocker2000explaining} logic, the recommender system finds a group of similar users similar as explanations. On the other hand, \textit{item-based collaborative filtering}~\cite{sarwar2001item} logic explains the recommendation by similar items. In addition, some explanations are based on user reviews. For example, prior works use topic modeling to generate word clouds as explanations ~\cite{zhang2014explicit,wu2015flame} and Costa~\etal~\cite{costa2018automatic} train language models on user review corpus to generate explanations. In addition, other works base explanations on social connections. For example, Sharma and Cosley show users a number of friends who also liked the item~\cite{sharma2013social}. For more details, refer to ~\cite{zhang2018explainable} for a general survey. Note that different explaining logic would require different explaining user interfaces, and therefore it is difficult to compare them in a controlled experimental setting. In this work, we focus on \textit{item-based collaborative filtering} logic only.
% For example, collaborative filtering logic displays the relevant users or items, user opinion based logic shows a text block, and social interaction based logic has different user interface designs, ranging from a text sentence/paragraph to links to source influence. 
% Because of their difference in displaying forms, it is difficult to compare explanations generated based on different logic in a controlled experimental setting. In this work, we focus on \textit{item-based collaborative filtering} logic.

\para{Counterfactual Explanation.} In classification task, Wachter~\etal~\cite{wachter2017counterfactual} use white-box optimization to find the smallest change on an input's features that can alter the model prediction. Mothilal~\etal~\cite{mothilal2020explaining} add a diversity loss into the optimization that maximizes the diversity between explanations. 
% Note that our definition of counterfactual examples is radically different from the conventional definition in classification. The classification definition assumes a \textit{fixed trained} model while our model is changeable with respect to the change of its training data because in our case the change in user history (\ie training data) impacts the model while in the classification case the change only happens in the test data.
% Russell~\etal~\cite{russell2019efficient} propose a mixed integer programming based approach to address the difficulty of generating sensible explanations regarding categorical features. 
Relatively fewer works consider counterfactual explanations in recommender system. Ghazimatin~\etal~\cite{ghazimatin2020prince} consider graph recommender models that encode user-item interactions. They propose a Personal PageRank based approach that searches for the minimal change on the graph. However, this approach is only applicable to PageRank-based recommendation models. In addition, Tran~\etal~\cite{tran2021counterfactual} propose an influence function based approach that estimates the influence of a training sample on model predictions.
% They use the influence function to compute how the predicted score on the explained item might change if some training items were removed. 
Furthermore, Kaffes~\etal~\cite{kaffes2021model} propose a black-box solution that performs Breadth First Search with heuristics that combine search length and drop of item rank.

\para{Predicting Model Predictions.} The essential goal of the surrogate model is to predict the recommendation model's predictions. There are a few works that start to study predicting model predictions recently. Ilya~\etal~\cite{ilyas2022datamodels} use a linear surrogate model to predict a neural network's predictions when a subset of the neural network's training samples are removed. They also find the simple linear model works well and the reason is unclear. 
Saunshi~\etal~\cite{saunshi2022understanding} suggest a linear model works might be because it can approximate functions related to test error well. 
In addition, our work differs fundamentally from instance-level explaining techniques using local surrogate models like LIME~\cite{ribeiro2016should} and SHAP~\cite{lundberg2017unified}. They focus on a \textit{fixed trained} model without considering the counterfactual impact of changing the training samples.

\para{Machine Unlearning.} Machine unlearning studies how to delete training samples from a model without retraining the model from scratch, which is related to how to quickly obtain the counterfactual model. Bourtoule~\etal~\cite{bourtoule2021machine} propose an exact unlearning framework, which replaces the original model with an ensemble model with each sub-model is trained on a data subset. Then removing a training sample is done by discarding its sub-model. In addition, Golatkar~\etal~\cite{golatkar2020eternal} use information theory methods to remove information from the trained model weights. Furthermore, Golatkar~\etal~\cite{golatkar2021mixed} assume there is a core data subset that is not removed, and train a backbone model on it. Then the model for the non-core data is a linear approximation model which is fast to unlearn. Researchers also propose to use caching information during training~\cite{wu2020deltagrad}, or design unlearning methods for a particular class of models like tree-based~\cite{brophy2021machine}.

\section{Conclusion}
\label{sec:conclusion}
We propose a learning-based framework to generate counterfactual explanations in a recommender system. Our key insight is by training a surrogate model to predict the effect on the recommendation caused by deleting a history subset, we can estimate the counterfactual impact that a deletion would have on a recommendation. And then we can generate the counterfactual explanations by searching for the deletion that is most likely to remove the recommended item from the recommendation list. We hope our work can inspire more research on counterfactual explanations, which is appealing as an alternative way to explain recommendations because it provides an explicitly stated and verifiable logic as well as greater user control.

% We point out two limitations and future work.  \textit{First}, in \textit{Simulating Counterfactual Outcomes} step, although we can parallelize retraining, we still need to (re)train a large number of models to generate enough ground-truth data in order to train surrogate models well. One future work is to accelerate this step by inexactly approximating retraining. \textit{Second}, we do not have a clear understanding of why linear surrogate models can generalize well when predicting model predictions, which is an open problem also pointed out by~\cite{ilyas2022datamodels}. 
% One future work is to test surrogate models that are middle grounds between linear and MLP model, \eg kernel ridge regression.

%%
%% The next two lines define the bibliography style to be used, and
%% the bibliography file.
\bibliography{references}
\bibliographystyle{unsrt}

%%
%% If your work has an appendix, this is the place to put it.
% \appendix
\appendix
% \section{MLP Surrogate Model Architecture}
% \label{app:mlp}
% TBD.

\section{Naive Approach of Generating Counterfactual Explanations}
\label{app:naive}

Algorithm~\ref{algo:naive} shows the forbiddingly slow naive approach of generating counterfactual explanations.

\section{Additonal Experiment Details}
\subsection{Dataset Details}
\label{app:data}

Table~\ref{tab:exp} shows the details of datasets used in our experiments.

\subsection{Full Experimental Results}
\label{app:full_exp}
Table~\ref{tab:mf_full} shows explanation performance, \ie percentage of top-K (K=1,3,5) recommended items are truly removed from the recommendation if we actually delete the claimed explanatory items and retrain the model, on MovieLens model. Table~\ref{tab:ncf_full} shows the same results on the Neural Collaborative Filtering model. The observations are largely consistent with descriptions in Section~\ref{subsec:results}.

\subsection{Performance Analysis of Baselines}
\label{app:baseline}

\begin{figure}[!t]
 \centering
    \subfloat[\centering Item embedding distance. ]{{\includegraphics[width=0.45\linewidth]{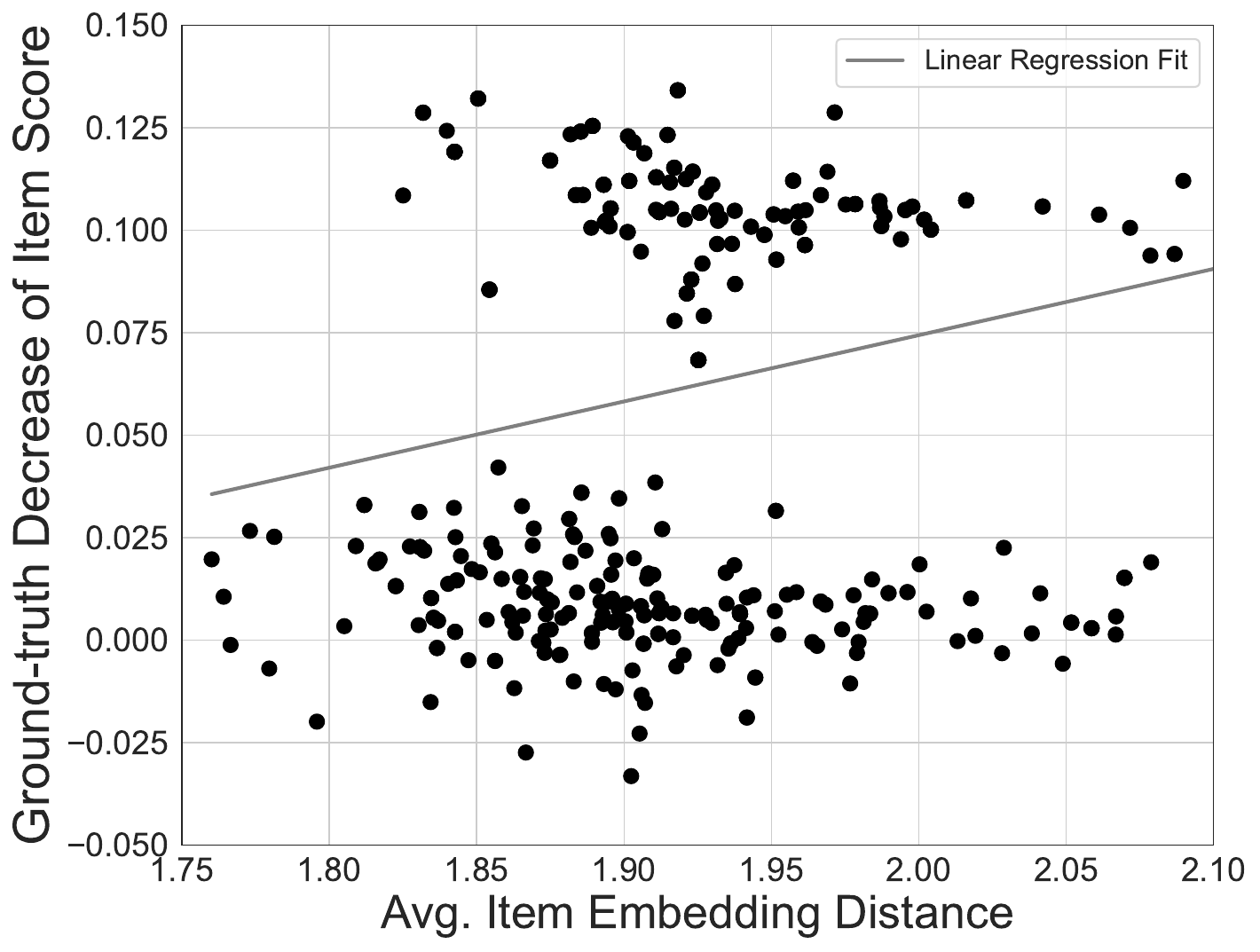} }}%
    \hfill
    \subfloat[\centering Influence function distance.]{{\includegraphics[width=0.45\linewidth]{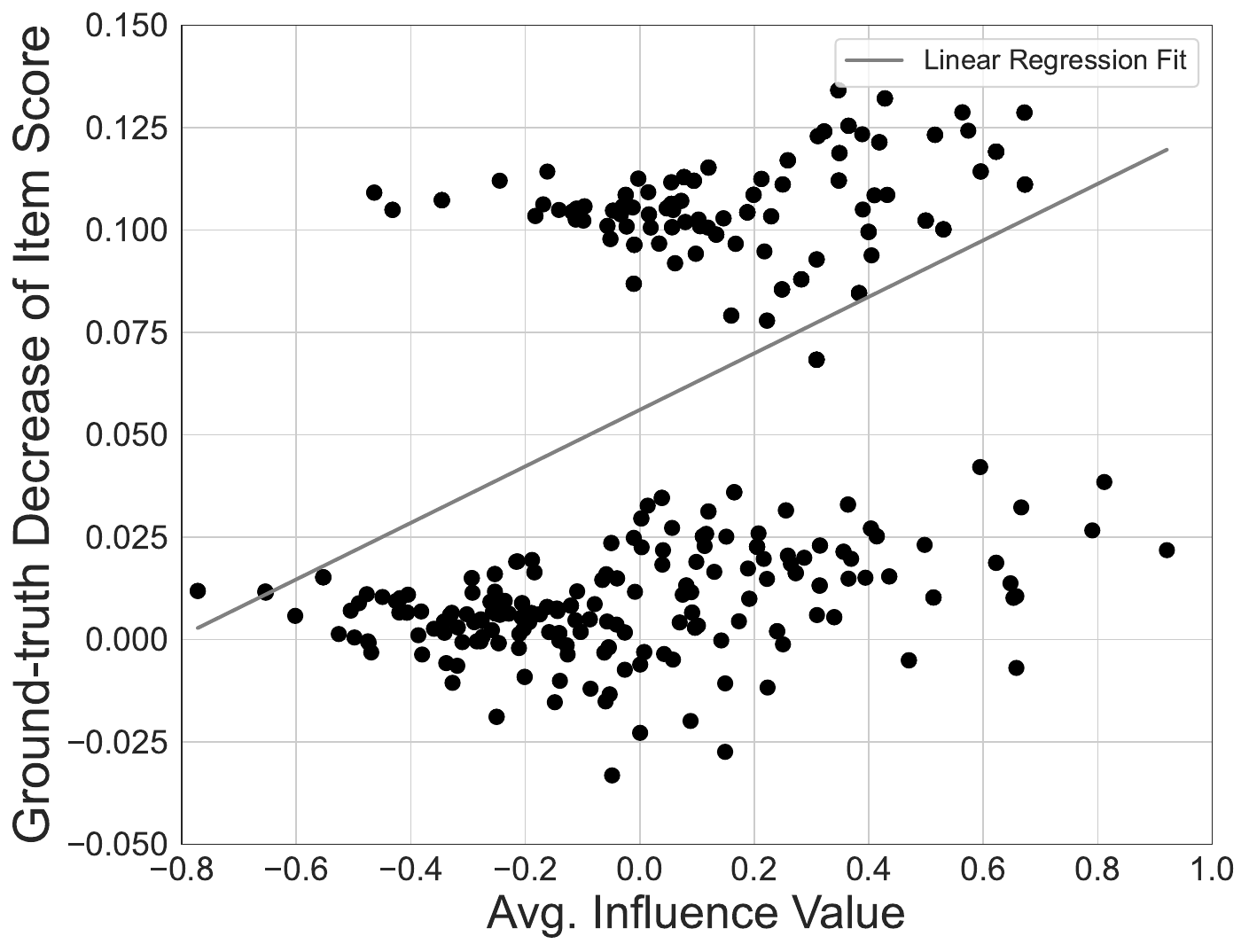} }}%
    \caption{(a): Average item embedding distance between an explanation and the recommended item vs. the ground-truth decrease of the recommended item's score if the explanation were actually removed and the model were retrained. (b): Average influence function value of an explanation vs. the ground-truth score decrease. Overall, neither distance measure is highly correlated with the ground-truth score decrease with Pearson correlation $0.175$ and $0.356$ only for item embedding distance and influence function respectively.}%
    \label{fig:dist}
\end{figure}

In Figure~\ref{fig:dist}, we show the relationship between the distance measured by KNN and Influence Function on explanations and the ground-truth score decrease of the recommendation if we actually delete explanations from the training set and retrain the model. As we can see, neither KNN nor Influence Function is highly correlated with the ground-truth score decrease with Pearson correlation $0.175$ and $0.356$ only.

\subsection{Ablation Study: Definition of $\Delta \hat{y}$}
\label{app:score}
\para{Rank Difference Definition.} Define $\pi_{u, \hat{\theta}}(i)$ to be the rank of the target item $i$ in the ranking list produced by the original model $\hat{\theta}$ and $\pi_{u, \hat{\theta}(-D_u)}(i)$ to be the item $i$'s rank in the counterfactual ranking list, then the rank difference definition of $\Delta \hat{y}$ is:
\begin{equation}
\label{eq:def_rank}
\Delta \hat{y}_{rank}(u, D_u, i) \vcentcolon= \pi_{u, \hat{\theta}(-D_u)}(i) - \pi_{u, \hat{\theta}}(i)
\end{equation}

If it is positive, it means item $i$ is ranked lower counterfactually than the original.

\para{Cross Ranking List Definition.} Recall that $f(u, i; \hat{\theta})$ is the predicted score on the target item $i$ by the original model and $\pi_{u, \hat{\theta}}^{-1}(1)$ is the top-1 item in the original ranking list; $f(u, i; \hat{\theta}(-D_u))$ is the predicted score on the target item $i$ by the counterfactual model and $\pi_{u, \hat{\theta}(-D_u)}^{-1}(1)$ is the top-1 item in the counterfactual ranking list, the cross ranking list definition of $\Delta \hat{y}$ is:
\begin{equation}
\label{eq:cross_rank}
% \resizebox{0.9\linewidth}{!}{%
\Delta \hat{y}_{cross}(u, D_u, i) \vcentcolon= \frac{f(u, i; \hat{\theta})}{f(u, \pi_{u, \hat{\theta}}^{-1}(1); \hat{\theta})} - \frac{f(u, i; \hat{\theta}(-D_u))}{f(u, \pi_{u, \hat{\theta}(-D_u)}^{-1}(1); \hat{\theta}(-D_u))}
% }
\end{equation}

If it is positive, it means the normalized score in the counterfactual list of item $i$ is smaller than the original, and therefore it is likely that the position will be lower in the counterfactual case.

\subsection{Ablation Study: Visualization of Surrogate Model's Training Inputs}
\label{app:vis}

\begin{figure}[!t]
  \centering
  \includegraphics[width=0.5\linewidth]{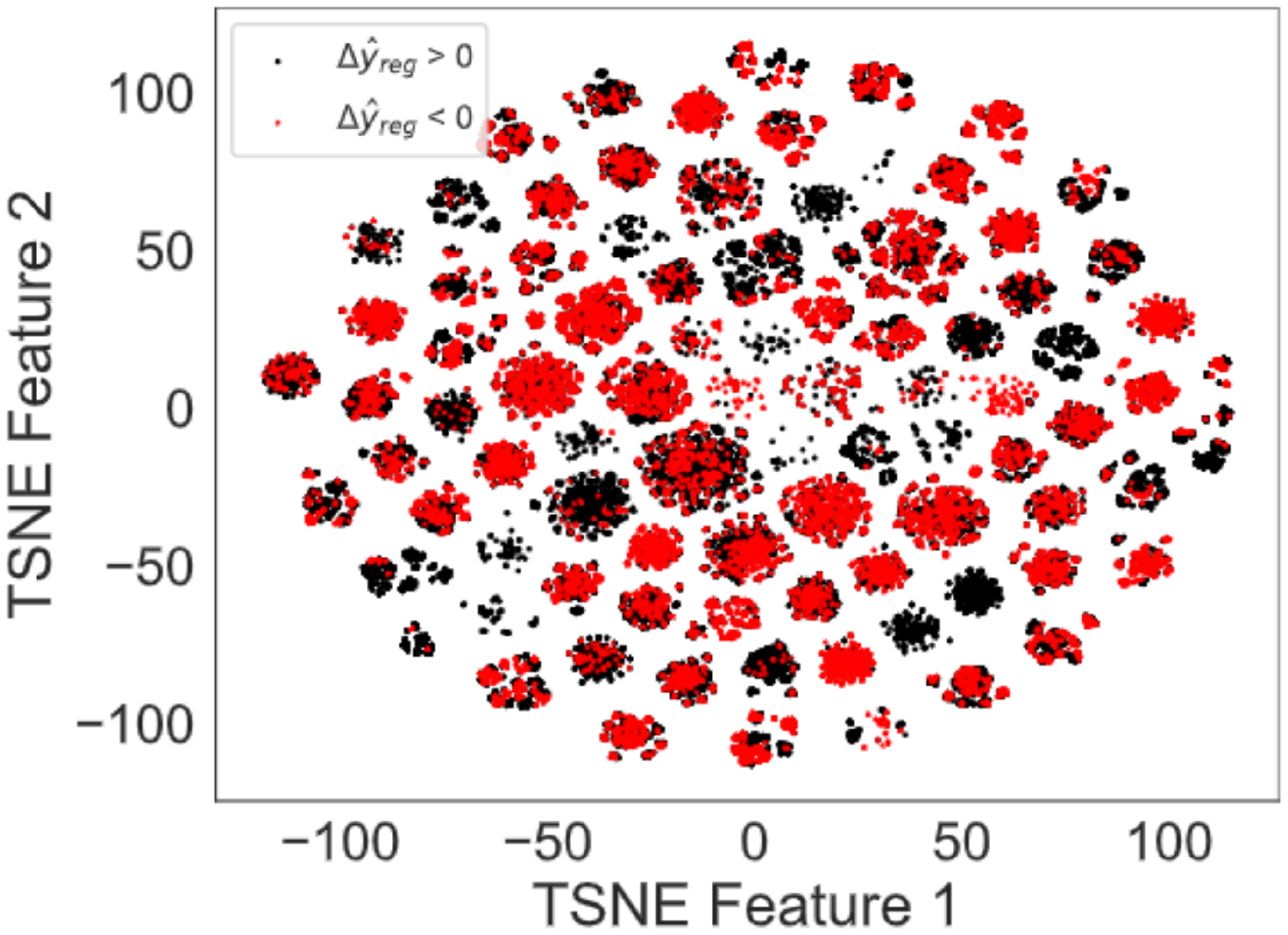}
  \caption{t-SNE visualization of surrogate model's training data (MovieLens data and Matrix Factorization model, with $\Delta \hat{y}_{reg}$ definition, on 100 training users) in 2-D space.}
  \label{fig:vis}
\end{figure}

In Figure~\ref{fig:vis}, we show t-SNE visualization of the training data of the surrogate model (on MovieLens data with Matrix Factorization model, with $\Delta \hat{y}_{reg}$ definition) in 2-D space. One hypothesis is because the recommendation model is Matrix Factorization, a (bi)linear model, its input-output relationship is linear as well, and therefore easier for a linear model to learn. However from the visualization, we do not observe a clear linear separability of the surrogate model's training data.

% \section{User Study: Details}
% \label{app:user}
% We include more details of our user study regarding quality control and results collection.

\begin{algorithm}[t]
\begin{algorithmic}[1]
\State \textbf{Input:} Target user: $u$, Target item: $i$, Rank Threshold: $K$.
\State \textbf{Output:} A counterfactual explanation $E_{u, i}$.
\For{$D_u \in 2^{I_u}$}
	\State Retrain to get counterfactual model $\hat{\theta}(- D_u)$ with eq (\ref{eq:cf_model}).
	\If{$\pi_{u, \hat{\theta}(- D_u)}(i) > K$}
	    \State Return $D_u$.
    \EndIf
\EndFor
\State Return $\emptyset$. 
\end{algorithmic}
\caption{Naive algorithm to generate counterfactual explanations.}
\label{algo:naive}
\end{algorithm}

\begin{table}[!t]
\centering
\resizebox{0.6\linewidth}{!}{
\begin{tabular}{@{}llll@{}}
\toprule
 & \# of Users & \# of Items & \# of Interactions \\ \midrule
MovieLens-100K~\cite{mvledata} & 943 & 1682 & 100,000 \\
Netflix (Small)~\cite{netflix} & 10,000 & 5,000 & 607,803 \\
Amazon Music~\cite{netflix} & 5,541 & 3,568 & 64,706 \\ \bottomrule
\end{tabular}
}
\caption{Datasets used in our experiments.}
% \vspace{-0.9cm}
\label{tab:exp}
\end{table}

\begin{table*}[!t]
\centering
\resizebox{0.75\linewidth}{!}{
\begin{tabular}{|l|lll|lll|lll|}
\hline
\multirow{2}{*}{} & \multicolumn{3}{c|}{MovieLens} & \multicolumn{3}{c|}{Netflix} & \multicolumn{3}{c|}{Amazon} \\ \cline{2-10} 
 & \multicolumn{1}{l|}{K = 1} & \multicolumn{1}{l|}{K = 3} & K = 5 & \multicolumn{1}{l|}{K = 1} & \multicolumn{1}{l|}{K = 3} & K = 5 & \multicolumn{1}{l|}{K = 1} & \multicolumn{1}{l|}{K = 3} & K = 5 \\ \hline
KNN & \multicolumn{1}{l|}{17\%} & \multicolumn{1}{l|}{10\%} & 3\% & \multicolumn{1}{l|}{30\%} & \multicolumn{1}{l|}{18\%} & 12\% & \multicolumn{1}{l|}{18.5\%} & \multicolumn{1}{l|}{8\%} & 5.5\% \\ \hline
Influence Function & \multicolumn{1}{l|}{35.5\%} & \multicolumn{1}{l|}{15.5\%} & 9.5\% & \multicolumn{1}{l|}{37\%} & \multicolumn{1}{l|}{18\%} & 10.5\% & \multicolumn{1}{l|}{37.5\%} & \multicolumn{1}{l|}{19.5\%} & 11.5\% \\ \hline
Linear Surrogate (reg) & \multicolumn{1}{l|}{\textbf{82\%}} & \multicolumn{1}{l|}{\textbf{68\%}} & \textbf{54.5\%} & \multicolumn{1}{l|}{\textbf{89\%}} & \multicolumn{1}{l|}{\textbf{73.5\%}} & \textbf{62\%} & \multicolumn{1}{l|}{\textbf{81\%}} & \multicolumn{1}{l|}{\textbf{57\%}} & \textbf{44\%} \\ \hline
MLP Surrogate (reg) & \multicolumn{1}{l|}{76.5\%} & \multicolumn{1}{l|}{52.5\%} & 42.5\% & \multicolumn{1}{l|}{80.5\%} & \multicolumn{1}{l|}{50.5\%} & 40.5\% & \multicolumn{1}{l|}{77\%} & \multicolumn{1}{l|}{54\%} & \textbf{44\%} \\ \hline
Linear Surrogate (clf) & \multicolumn{1}{l|}{79\%} & \multicolumn{1}{l|}{61\%} & 51\% & \multicolumn{1}{l|}{85\%} & \multicolumn{1}{l|}{68.5\%} & 57.5\% & \multicolumn{1}{l|}{71\%} & \multicolumn{1}{l|}{50.5\%} & 40.5\% \\ \hline
MLP Surrogate (clf) & \multicolumn{1}{l|}{72.5\%} & \multicolumn{1}{l|}{48\%} & 41\% & \multicolumn{1}{l|}{76\%} & \multicolumn{1}{l|}{61\%} & 51\% & \multicolumn{1}{l|}{68\%} & \multicolumn{1}{l|}{48.5\%} & 38.5\% \\ \hline
\end{tabular}
}
\caption{Full results on Matrix Factorization model. Percentage of recommended items that are removed from the top-K (K=1,3,5) recommendation list if the generated explanations were actually deleted and the model were retrained.
}
% \vspace{-0.3cm}
\label{tab:mf_full}
\end{table*}

\begin{table*}[!t]
\centering
\resizebox{0.75\linewidth}{!}{
\begin{tabular}{|l|lll|lll|lll|}
\hline
\multirow{2}{*}{} & \multicolumn{3}{c|}{MovieLens} & \multicolumn{3}{c|}{Netflix} & \multicolumn{3}{c|}{Amazon} \\ \cline{2-10} 
 & \multicolumn{1}{l|}{K = 1} & \multicolumn{1}{l|}{K = 3} & K = 5 & \multicolumn{1}{l|}{K = 1} & \multicolumn{1}{l|}{K = 3} & K = 5 & \multicolumn{1}{l|}{K = 1} & \multicolumn{1}{l|}{K = 3} & K = 5 \\ \hline
KNN & \multicolumn{1}{l|}{42\%} & \multicolumn{1}{l|}{27\%} & 21.5\% & \multicolumn{1}{l|}{38.5\%} & \multicolumn{1}{l|}{26\%} & 24\% & \multicolumn{1}{l|}{40\%} & \multicolumn{1}{l|}{28\%} & 21\% \\ \hline
Influence Function & \multicolumn{1}{l|}{44\%} & \multicolumn{1}{l|}{29.5\%} & 23.5\% & \multicolumn{1}{l|}{40.5\%} & \multicolumn{1}{l|}{28.5\%} & 24\% & \multicolumn{1}{l|}{41\%} & \multicolumn{1}{l|}{29\%} & 23\% \\ \hline
Linear Surrogate (reg) & \multicolumn{1}{l|}{\textbf{86\%}} & \multicolumn{1}{l|}{\textbf{60\%}} & \textbf{44\%} & \multicolumn{1}{l|}{\textbf{68.5\%}} & \multicolumn{1}{l|}{\textbf{41.5\%}} & \textbf{32\%} & \multicolumn{1}{l|}{\textbf{74.5\%}} & \multicolumn{1}{l|}{\textbf{54\%}} & \textbf{40.5\%} \\ \hline
MLP Surrogate (reg) & \multicolumn{1}{l|}{83\%} & \multicolumn{1}{l|}{57.5\%} & 42\% & \multicolumn{1}{l|}{\textbf{68.5\%}} & \multicolumn{1}{l|}{39.5\%} & 30\% & \multicolumn{1}{l|}{71\%} & \multicolumn{1}{l|}{52\%} & 38.5\% \\ \hline
Linear Surrogate (clf) & \multicolumn{1}{l|}{81\%} & \multicolumn{1}{l|}{53.5\%} & 38.5\% & \multicolumn{1}{l|}{61.5\%} & \multicolumn{1}{l|}{36.5\%} & \textbf{32\%} & \multicolumn{1}{l|}{72\%} & \multicolumn{1}{l|}{49.5\%} & 38.5\% \\ \hline
MLP Surrogate (clf) & \multicolumn{1}{l|}{75.5\%} & \multicolumn{1}{l|}{49.5\%} & 36.5\% & \multicolumn{1}{l|}{63.5\%} & \multicolumn{1}{l|}{35\%} & 27.5\% & \multicolumn{1}{l|}{68.5\%} & \multicolumn{1}{l|}{48.5\%} & 36\% \\ \hline
\end{tabular}
}
\caption{Full results on Neural Collaborative Filtering model. Percentage of recommended items that are removed from the top-K (K=1,3,5) recommendation list if the generated explanations were actually deleted and the model were retrained.
}
% \vspace{-0.3cm}
\label{tab:ncf_full}
\end{table*}

\end{document}